\documentclass[conference]{IEEEtran}
\IEEEoverridecommandlockouts

\usepackage{amsmath}
\usepackage{algorithm,algorithmic}
\usepackage{graphicx}

\makeatletter
\let\MYcaption\@makecaption
\makeatother

\usepackage[font=footnotesize]{subcaption}

\makeatletter
\let\@makecaption\MYcaption
\makeatother

\usepackage[style=ieee,maxbibnames=3]{biblatex}
\begin{document}

\title{Mastering the Game of Sungka from Random Play\\
}

\author{\IEEEauthorblockN{Raimarc Dionido\textsuperscript{*}\thanks{\textsuperscript{*}Authors contributed equally} and Darwin Bautista\textsuperscript{*}}
\IEEEauthorblockA{\textit{Electrical and Electronics Engineering Institute} \\
\textit{University of the Philippines Diliman}\\
Quezon City, Philippines \\
\texttt{\{raimarc.dionido,darwin.bautista\}@eee.upd.edu.ph}
}
}

\maketitle

\begin{abstract}
Recent work in reinforcement learning demonstrated that learning solely through self-play is not only possible, but could also result in novel strategies that humans never would have thought of. However, optimization methods cast as a game between two players require careful tuning to prevent suboptimal results. Hence, we look at random play as an alternative method.

In this paper, we train a DQN agent to play \textit{Sungka}, a two-player turn-based board game wherein the players compete to obtain more stones than the other. We show that even with purely random play, our training algorithm converges very fast and is stable. Moreover, we test our trained agent against several baselines and show its ability to consistently win against these.
\end{abstract}

\begin{IEEEkeywords}
reinforcement learning, DQN, Sungka
\end{IEEEkeywords}

\section{Introduction}

Recent progress in deep learning has fueled breakthroughs in reinforcement learning \cite{mnih2015human}. However, most successful works rely on annotated data which are expensive to acquire and labor-intensive to prepare. Deep reinforcement learning models have been of increasing interest due to the ability to learn from own experiences. It has been used to allow an agent to solve and play various games such as Atari games and board games such as Go and Chess.

AlphaGo \cite{AlphaGo} was able to achieve superhuman performance in the game of Go and was able to defeat champions of the game. They were able to achieve this by training it with supervision, then training it as a reinforcement learning problem. This was further improved in AlphaGoZero \cite{alphagozero} which learned to play Go through solely reinforcement learning, without any human data supervision. This suggests that learning a game solely through self-play is not only possible, but could also result in novel strategies that no human would ever have thought of.

Motivated by the results of AlphaGoZero, we seek to further explore other means of learning. Self-play is a good learning mechanism assuming that the agent gets better over time. However, this requires careful optimization to prevent suboptimal results often seen in optimization methods cast as a game between two agents \cite{arjovsky2017wasserstein}. Particularly, we turn to purely random play as an alternative optimization method and develop a fast-converging algorithm around it. To test our hypothesis, we choose the game of Sungka, a Filipino variant of Mancala. It is a two-player turn-based board game wherein each player tries to collect as many stones as they can.

The game looks deceptively simple because the Sungka board only has 2 heads, 14 houses, and 98 stones in total. However, the actual state-space complexity is $|S| = C_{16}^{(98)} = 1.81\times{10}^{18}$
which makes it more complex than various games such as Nine Men's Morris, Connect Four, Pentominoes, and Domineering, and comparable to American Checkers as seen in Table \ref{tab:state-space}. This level of complexity makes Sungka a good candidate for experimenting with random play as a mechanism for learning.

\begin{table}
    \caption{State-space complexities of various games}
    \label{tab:state-space}
    \begin{center}
     \begin{tabular}{|c|c|} 
     \hline
      Game & $|S|$ ($\log_{10}$)\\
     \hline
     \hline
      Nine Men's Morris & 10 \cite{allis1994searching} \\
      \hline
      Pentominoes & 12 \cite{van2002games}\\
      \hline
      Connect Four & 13 \cite{allis1994searching} \\
      \hline
      Domineering & 15 \cite{van2002games} \\
      \hline
      American Checkers & 18 \cite{allis1994searching} \\
      \hline
      \textbf{Sungka} & \textbf{18} \\
      \hline
    \end{tabular}
    \end{center}
\end{table}

In this paper, we present a reinforcement learning agent capable of playing Sungka at human-level performance. We also show empirical evidence that with just random play, our training algorithm still converges fast, and that the trained agent discovers various strategies such as maximizing the number of consecutive turns, and choosing an action which would result in \textit{sunog}.

Specifically, our contributions\footnote{Source code at: \texttt{https://github.com/baudm/sungka-ai}} are as follows:
\begin{enumerate}
    \item OpenAI Gym environment for Sungka
    \item Reward formulation which penalizes actions resulting in high opponent scores
    \item Fast-converging and stable training algorithm 
\end{enumerate}

\begin{figure}[htbp]
\centerline{\includegraphics[scale=0.4]{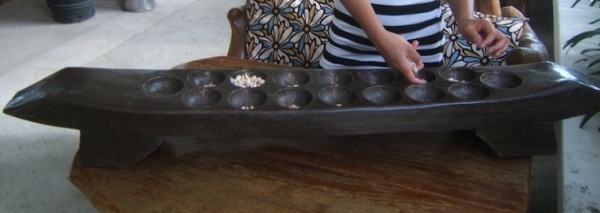}}
\caption{Image of a Sungkahan, the board used in playing sungka \cite{sungkaboard}}
\label{fig}
\end{figure}

\section{Related Work}

As far as we know, there is no prior work yet on Sungka, but there have been several works on
Mancala and its other variants.

Prior to Deep Q Networks (DQN) \cite{mnih2015human}, the successful use of deep neural networks in the context of reinforcement learning has not yet been demonstrated. An early attempt at using neural nets for developing an intelligent agent for Congkak \cite{chepa2013application}, one of the Malaysian traditional games and a variant of Mancala, was largely unsuccessful. Results showed that the neural net policy is even worse than a random policy.

Pinto et al. \cite{PintoMancala} used three agents (Game trees, Q-learning, Rule of Thumb) and six reward functions.
Results show that Q-learning beats mini-max, and that RoT is efficient but is easily beaten. The reward function has a bigger impact on the game outcome than the type of agent used. DaVolio and Langenborg \cite{DaVolioAI} compared eight agents (Random, Max, Exact, MinMax, MCTS, Q-learning, Deep Q, A3C) and found that the agent performances were roughly in line with complexity: Random Agent as the worst and A3C Agent was the best.

\section{Methodology}

\subsection{Game Mechanics}
\begin{figure}[htbp]
    \centerline{\includegraphics[scale=0.6]{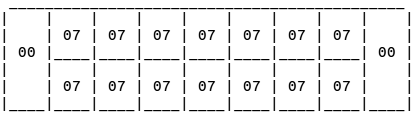}}
    \caption{Initial state of the sungka board. There are seven houses each on each player's side. Each player also has his own head where the player on top has the right head while the player on the bottom has the left head.}
    \label{fig:sungkarender}
\end{figure}

Sungka is a two-player board game where each player takes turn in moving stones with the objective of obtaining the most stones in their respective \textit{heads}. Each player has seven \textit{houses} each filled with seven stones initially as shown in Figure \ref{fig:sungkarender}.

A player chooses one of the houses on his side, takes the stones in it, and then moves in a clockwise direction while dropping one stone on each house or head he passes over excluding the other player's head. If the last stone is dropped into his own head, the player continues his turn by choosing any of houses on his side again. If the last stone is dropped on any filled house, the player picks up all the shells in that house (including the last stone dropped) and continue the turn.

A player's turn ends when the last stone is dropped into an empty house. If this empty house is on the player's side, this player takes all the stones on the other player's house directly opposite of it and the last stone dropped itself and puts it in his own head. This mechanic is called \textit{sunog}. Otherwise, the other player will now choose his move. The game ends when all stones are dropped into any of the heads. The player with more stones in his head wins the game.

\subsection{Environment Limitations}

In an actual game of Sungka, both players start their first move simultaneously. When a player has finished his move, he must wait until the other player is finished. When both players have finished their first move, the players make actions in alternating fashion where the player who finished his first move faster moves first. This gives the game a real-time element to it. For this project, we omit this mechanism and limit the game to a purely turn-based game. 

After the first round of Sungka, each player redistributes their collected stones back into their houses with seven stones each. If a house could not be filled with exactly seven stones, the house is not filled and is burnt.  After filling each house, all excess stones are returned to their head. The game is played again while each player no longer putting stones in the burnt houses, with the winner of the previous round moving first. The game ends when a player had all his houses burnt or surrenders. For simplicity, we only play a single round of sungka and declare the player with more stones after this round as the winner.

\subsection{Sungka Environment}
We implement our environment using the OpenAI Gym toolkit \cite{OpenAIGym}. Since the gym environment does not explicitly support turn-based games, we manually enforce the turn-based nature of the actions by keeping track of the players' turns.

\subsubsection{Observation Space}
An observation is a representation of the game board state. It is a 1x14 array where the first seven elements represent the first player's side, while the last seven elements represent the second player's side. The value inside the array corresponds to the number of stones the house contains. The heads effectively contain the players' current scores. Since the current scores do not affect the decision-making process, we exclude them from the state vector.

\subsubsection{Action Space}
A player has seven possible actions which correspond to choosing one of the seven houses on his side of the board. In the actual environment, the action space is 14. We map the player-specific actions to the raw indices used in the environment: 0-6 for Player 1 and 7-13 for Player 2.

\subsubsection{Rewards}
An action made by a player does not contribute only to the player's own score, but also affects the maximum score attainable by the opponent in the next turn. Since the player with the most number of stones wins, the agent should not only maximize its own score, but should also minimize the opponent's score.

We define a timestep such that each one consists of two turns: the agent's and the opponent's. We denote $r_t$ as the reward for timestep $t$, and $r_{t,agent}$ and $r_{t,opponent}$ as the scores obtained by the agent and the opponent at timestep $t$, respectively. Thus we formulate the reward for each timestep as shown in \eqref{eqn:reward}.

\begin{equation}
\label{eqn:reward}
r_{t} = r_{t,agent} - r_{t,opponent}
\end{equation}

\subsection{Deep Q-Learning}\label{subsec:deep-q}
The game board consists of 2 heads, 14 houses, and 98 stones in total. We can model the board configuration as a combination with replacement problem. As such, the number of theoretically possible game states is $|S| = C_{16}^{(98)} = 1.81\times{10}^{18}$. With this large number of states, using a Q-table to store the values for all state-action pairs becomes impractical, if not impossible. Thus, we instead base our approach on DQN and use a neural network to learn the optimal Q function.

\subsection{Baselines}
We test our trained DQN agent against several policies:
\begin{enumerate}
    \item \textit{Random Policy:} The random policy agent simply chooses a random action from a uniform distribution.
    \item \textit{Max Policy:} The max policy agent always chooses the house with the most number of stones.
    \item \textit{Exact Policy:} The exact policy agent chooses the nearest house to the head where the number of stones is equal to its distance to the head. This allows the agent to get another turn. If there are more than one houses which satisfy the condition, the nearer house to the head is chosen first. If no house satisfies the condition, max policy is used.
    \item \textit{DQN Agent:} The trained DQN agent plays against itself.
\end{enumerate}

\section{Experiments}

We generate training episodes by making the DQN Agent play against the Random Agent. We train the DQN every step of an episode, for a total of 10,000 episodes. We employ Experience Replay with a buffer size of 2,000, and sample a random mini-batch of size 128 every training iteration. Algorithm \ref{alg:training} describes the training procedure.

\begin{algorithm}
\caption{Turn-based DQN Training Procedure}
\begin{algorithmic}[1]
\label{alg:training}
\REQUIRE $N$ (number of episodes) $\epsilon$, $\gamma$ (discount factor)
\ENSURE Trained DQN model
\STATE Initialize $\theta$ parameters of model
\FOR{$episode=0$ \TO $N$}
    \STATE Initialize $S$
    \REPEAT
        \STATE Choose $A$ from $S$ using $\epsilon$-greedy DQN policy
        \STATE Take action $A$, observe $R$, $S'$, $P$
        \WHILE{$P$ = Opponent}
            \STATE Opponent chooses $A_{opp}$ from $S'$
            \STATE Take action $A_{opp}$, observe $R_{opp}$, $S'$, $P$
            \STATE $R \xleftarrow{} R - R_{opp}$
        \ENDWHILE
        \STATE Store $S$, $A$, $R$, $S'$ in Experience Buffer
        \STATE $S \xleftarrow{} S'$
        \STATE \COMMENT{In a subroutine:}
        \STATE Sample $S$, $A$, $R$, $S'$ batch from Experience Buffer
        \IF{$episode\bmod{100} = 0$}
            \STATE $\theta_{target} \xleftarrow{} \theta$
        \ENDIF
        \STATE $V \xleftarrow{} Q(S, A; \theta)$
        \STATE $V_{target} \xleftarrow{} R + \gamma \max_{a} Q(S', a; \theta_{target})$
        \STATE $L \xleftarrow{} MSE(V, V_{target})$
        \STATE Update $\theta$ via gradient descent
    \UNTIL{episode ends}
\ENDFOR
\end{algorithmic}
\end{algorithm}

We explored various training setups, but we highlight our experiences in two scenarios:
\subsubsection{Annealed $\epsilon$}
This is the typical approach used in most reinforcement learning work because it ensures that in the early phases of training, the DQN gets trained on a very varied set of states. However, we found that for Sungka, starting with $\epsilon = 0.9$ and annealing it to $\epsilon = 0.05$ resulted in relatively \textit{unstable} training.

\subsubsection{Fixed $\epsilon$}
Using a fixed $\epsilon = 0.05$ resulted in the fastest convergence and most stable training. While it seems counterintuitive, note that the practical state space is smaller than the theoretical maximum, and that the stochastic behavior of the Random Agent opponent already provides ample \textit{exploration}.

Regardless of the two approaches, the DQN Agent always achieves human-level performance at the end of training.

At test time, we initially used full exploitation mode by setting epsilon to zero. However, when the model is still not trained well enough, the agent gets stuck on choosing an action which does not have any stones in it. To prevent this, we use a very small epsilon instead of using zero during test time.

\section{Results}

\begin{figure}[htbp]
    \centerline{\includegraphics[scale=0.6]{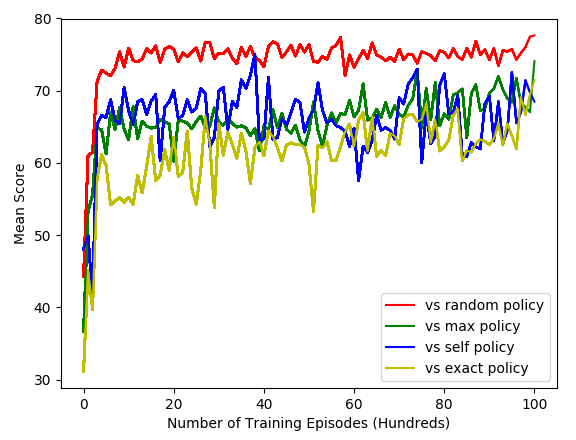}}
    \caption{Mean score versus various opponent agents averaged for 100 test episodes. Each mean score data point is calculated after every 100 training episodes.}
    \label{fig:rewards-progress}
\end{figure}

\begin{figure}[htbp]
    \centerline{\includegraphics[scale=0.6]{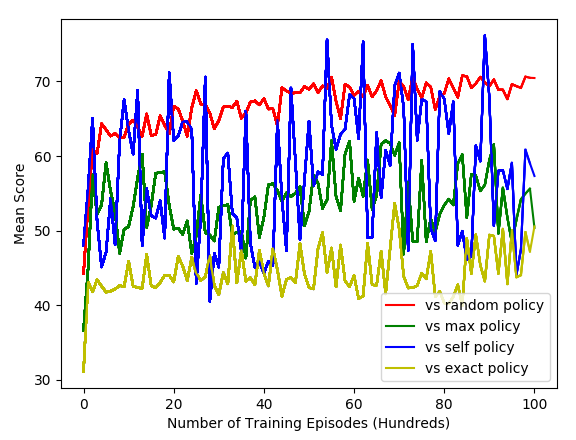}}
    \caption{Effect of reward formulation. Mean score versus various opponent agents averaged for 100 test episodes. Each data point is calculated after every 100 training episodes.}
    \label{fig:score-progress-p1-score-only}
\end{figure}

\begin{figure}[htbp]
    \centerline{\includegraphics[scale=0.6]{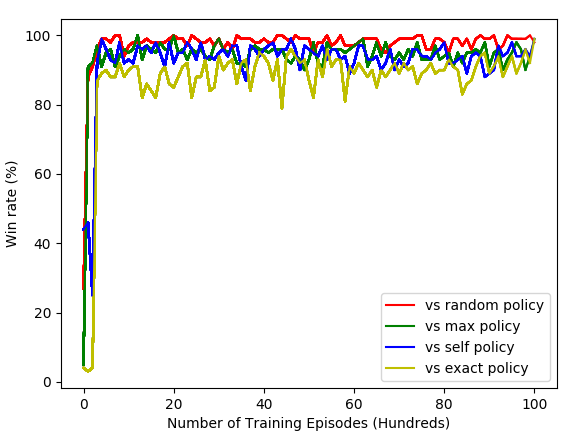}}
    \caption{Win rate versus various opponent agents averaged for 100 test episodes. Each data point is calculated after every 100 training episodes.}
    \label{fig:win-rate-progress}
\end{figure}

\begin{table}[htbp]
    \caption{Performance of Trained DQN Agent versus Other Policies over 1000 Test Episodes with $\epsilon=0.01$}
    
    \begin{subtable}[h]{0.45\textwidth}
    \centering
     \begin{tabular}{|c||c|c|} 
     \hline
     Policy & Average Reward & Win Percentage\\ [0.5ex] 
     \hline
     \hline
     vs. Random Policy & 77.621 & 99.4\% \\ 
     \hline
     vs. Max Policy & 74.122 & 99.5\% \\
     \hline
     vs. Exact Policy & 70.943 & 97.9\%  \\
     \hline
     vs. Self & 68.541 & 98.6\%  \\
     \hline
    \end{tabular}
    \caption{As First Turn}
    \label{tab:perfp1}
    \end{subtable}
    
    \begin{subtable}[h]{0.45\textwidth}
    \centering
     \begin{tabular}{|c||c|c|} 
     \hline
     Policy & Average Reward & Win Percentage\\ [0.5ex] 
     \hline
     \hline
     vs. Random Policy & 55.927 & 73.5\% \\ 
     \hline
     vs. Max Policy & 37 & 0.0\% \\
     \hline
     vs. Exact Policy & 29.995 & 0.0\%  \\
     \hline
     vs. Self & 29.459 & 1.1\%  \\
     \hline
    \end{tabular}
    \caption{As Second Turn}
    \label{tab:perfp2}
    \end{subtable}
    
\end{table}
\subsection{Effects of Reward Formulation}
Results show that training our DQN agent versus a random agent allows it to learn how to play sungka such that it maximizes the number of stones it can place in its own head. Figure \ref{fig:rewards-progress} shows that the agent is able to increase the average reward as training progresses. Moreover, the effect of the reward formulation in \eqref{eqn:reward} is apparent in Figure \ref{fig:score-progress-p1-score-only}. With everything held constant, the agent trained using the naive reward formulation, $r_t = r_{t,agent}$, is consistently outscored by the \textit{Exact} policy, and has an inconsistent performance against the \textit{Max} policy.

\subsection{Performance of First Turn Player vs. Second Turn Player}
During training, our DQN agent gets a high win percentage against any of the four policies tested after a few hundred episodes as shown in Figure \ref{fig:win-rate-progress}. Among the four policies, our agent had the hardest time against exact policy as our agent had the lowest win rate and the second lowest average reward when playing against it as shown in Table \ref{tab:perfp1}. This can be accounted by the ability of exact policy to maximize its number of turns which the other policies do not have. 

Table \ref{tab:perfp1} shows the performance of the final DQN agent aginst other policies over 1000 test episodes with $\epsilon = 0.01$. It can be seen that the trained agent performs very well against any policy when it has the first turn. When it plays as the second turn, it still wins against random policy at a good rate. However,  it could not win against max policy and exact policy. It also only wins 11 games out of 1000 versus itself. This shows that the game of sungka is biased towards the player that gets the first turn.

\begin{table}
    \caption{Performance of Trained Player2DQN Agent versus Other Policies over 1000 Test Episodes with $\epsilon=0.01$}
    \begin{subtable}[h]{0.45\textwidth}
    \centering
     \begin{tabular}{|c||c|c|} 
     \hline
     Policy & Average Reward & Win Percentage\\ [0.5ex] 
     \hline
     \hline
     vs. Random Policy & 65.271 & 94.5\% \\ 
     \hline
     vs. Max Policy & 40.871 & 0.2\% \\
     \hline
     vs. Exact Policy & 30.986 & 0.0\%  \\
     \hline
     vs. Self & 38.216 & 4.3\%  \\
     \hline
    \end{tabular}
    \caption{As Second Turn}
    \label{tab:perfp2trainedp2}
    \end{subtable}
    
    \begin{subtable}[h]{0.45\textwidth}
    \centering
     \begin{tabular}{|c||c|c|} 
     \hline
     Policy & Average Reward & Win Percentage\\ [0.5ex] 
     \hline
     \hline
     vs. Random Policy & 60.314 & 86.7\% \\ 
     \hline
     vs. Max Policy & 49.237 & 5.6\% \\
     \hline
     vs. Exact Policy & 51.838 & 96.3\%  \\
     \hline
     vs. Self & 59.784 & 95.5\%  \\
     \hline
    \end{tabular}
    \caption{As First Turn}
    \label{tab:perfp1trainedp2}
    \end{subtable}
\end{table}

Looking into this further, we trained another DQN agent which plays second (Player2DQN). Table \ref{tab:perfp2trainedp2} shows the performance of that agent when playing second. The average rewards received by Player2DQN are higher when playing against any of the policies than when Player1DQN plays as second turn. Win percentage against random policy and self is also higher than when it was not trained to play second.

We also look into the performance of Player2DQN as the first player in Table \ref{tab:perfp1trainedp2}. It shows that the average reward increased against max policy, exact policy, and self. Its win percentage against exact policy and self also drastically increased. 

Results from these experiments show that when playing against max policy and exact policy, having the first turn leads to more rewards and a higher win percentage. Player2DQN performed well against random, exact, and self when playing first even though it was trained as second. On the other hand, performance against random policy is dependent on how the agent was trained.

We also tested our Player1DQN agent against Player2DQN agent. Table \ref{tab:perfpvp} shows that the first player always has the advantage.

\begin{table}
    \caption{Performance of Trained Player1DQN Agent versus Player2DQN Agent}
    \label{tab:perfpvp}
    \begin{center}
     \begin{tabular}{|c||c|c|} 
     \hline
      & Average Reward & Win Percentage\\ [0.5ex] 
     \hline
     \hline
     Player1DQN as 1st turn& 71.991 & 98.1\% \\ 
     \hline
     Player2DQN as 2nd turn& 26.009 & 1.8\% \\
     \hline
     Draw & & 0.1\% \\
     \hline
     \hline
     Player2DQN as 1st turn& 68.189 & 98.0\% \\ 
     \hline
     Player1DQN as 2nd turn& 29.811 & 1.8\% \\
     \hline
     Draw & & 0.2\% \\
     \hline
    \end{tabular}
    \end{center}
\end{table}

\begin{figure}[htbp]
    \centerline{\includegraphics[scale=0.45]{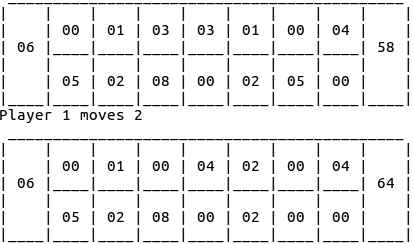}}
    \caption{Agent learns to maximize reward by choosing a move where the last stone ends on an empty house on his side which allows the agent to get the stones on the opponent's side.}
    \label{fig:sunog}
\end{figure}

\begin{figure}[htbp]
    \centerline{\includegraphics[scale=0.45]{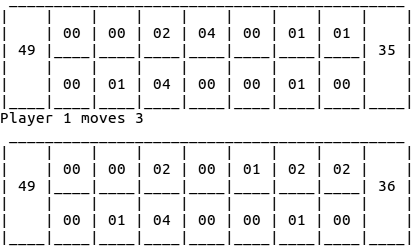}}
    \caption{Agent is able to maximize its move by choosing a move where the last stone ends on his own head. This allows the agent to choose another move which can further maximize the reward.}
    \label{fig:maximize-turn}
\end{figure}

\subsection{Learned Actions based on Game Mechanics}
Figures \ref{fig:sunog} and \ref{fig:maximize-turn} shows the agent's ability to exploit some of the game mechanics. In Figure \ref{fig:sunog} showed that the agent has learned to choose a move that will put the last stone on an empty house on his side of the board. This move allows him to get more stones since he also gets the stones on the opponent's side. In Figure \ref{fig:maximize-turn} exploits the fact that putting the last stone on his head allows him to make another move. This allows the agent to get one point, and gets to take another action without the board state changing unpredictably (due to an opponent's action).

\section{Conclusion and Recommendation}
We have trained a network that is capable of playing and winning in Sungka. The trained agent is able to choose actions that maximizes its reward to increase its probability of winning the game. We showed that the reward formulation which uses both the score accumulated by the other player and the agent's score for that turn result to more stable training and better performance. 

In this paper, we only trained a DQN agent. We recommend looking into the performance of other reinforcement learning methods such as cross entropy, trust region policy optimization, proximal policy optimization, and A3C. It would also be interesting to see the performance of agents trained using different reinforcement learning methods against each other.

\renewcommand*{\UrlFont}{\ttfamily}
\printbibliography

\end{document}